\documentclass{article}

\usepackage[preprint]{neurips_2024}

\usepackage[utf8]{inputenc} 
\usepackage[T1]{fontenc}    
\usepackage{hyperref}       
\usepackage{url}            
\usepackage{booktabs}       
\usepackage{amsfonts}       
\usepackage{nicefrac}       
\usepackage{microtype}      
\usepackage{xcolor}         

\usepackage{stfloats}
\usepackage{float}
\usepackage{subcaption}
\usepackage{amsmath}
\usepackage{amssymb}
\usepackage{mathtools}
\usepackage{amsthm}
\usepackage{algorithm}
\usepackage{algpseudocode}
\usepackage{authblk}

\usepackage[capitalize,noabbrev]{cleveref}

\title{Generative Pre-Trained Transformer for Symbolic Regression Base In-Context Reinforcement Learning}

\author[1]{Yanjie Li}
\author[1,2,*]{Weijun Li}
\author[1,*]{Lina Yu}
\author[1,*]{MinWu}
\author[1,2]{JingyiLiu}
\author[1,2]{WenqiangLi}
\author[1]{MeilanHao}
\author[1,2]{SuWei}
\author[1,2]{YusongDeng}

\affil[1]{ANNLAB, Institute of Semiconductors, Chinese Academy of Sciences, Beijing, China}
\affil[2]{School of Electronic, Electrical and Communication Engineering, University of Chinese Academy of Sciences, Beijing, China}

\affil[*]{Corresponding author: liyanjie@semi.ac.cn}

\begin{document}

\maketitle

\begin{abstract}
  The mathematical formula is the human language to describe nature and is the essence of scientific research. Therefore, finding mathematical formulas from observational data is a major demand of scientific research and a major challenge of artificial intelligence. This area is called symbolic regression. Originally symbolic regression was often formulated as a combinatorial optimization problem and solved using GP or reinforcement learning algorithms. These two kinds of algorithms have strong noise robustness ability and good Versatility. However, inference time usually takes a long time, so the search efficiency is relatively low. Later, based on large-scale pre-training data proposed, such methods use a large number of synthetic data points and expression pairs to train a Generative Pre-Trained Transformer(GPT). Then this GPT can only need to perform one forward propagation to obtain the results, the advantage is that the inference speed is very fast. However, its performance is very dependent on the training data and performs poorly on data outside the training set, which leads to poor noise robustness and Versatility of such methods. So, can we combine the advantages of the above two categories of SR algorithms? In this paper, we propose \textbf{FormulaGPT}, which trains a GPT using massive sparse reward learning histories of reinforcement learning-based SR algorithms as training data. After training, the SR algorithm based on reinforcement learning is distilled into a Transformer. When new test data comes, FormulaGPT can directly generate a "reinforcement learning process" and automatically update the learning policy in context.

Tested on more than ten datasets including SRBench, formulaGPT achieves the state-of-the-art performance in fitting ability compared with four baselines. In addition, it achieves satisfactory results in noise robustness, versatility, and inference efficiency.
\end{abstract}

\section{Introduction}
\label{submission}

As a bridge between human beings and the natural world, formulas play a pivotal role in the field of natural science. They not only condense abstract natural laws into precise mathematical language but also enable us to describe and analyze natural phenomena quantitatively. Symbolic regression, as a kind of data modeling method, aims to let the computer dig out the inherent mathematical laws from the observed data and reveal the hidden patterns and laws of the data. By discovering these underlying mathematical expressions, symbolic regression can reveal the nature of the interactions between variables and construct predictive models for future events. Specifically, if there is a set of data $[x_1,x_2,...,x_m,y]$ where $x_i\in \mathbb{R}^n$ and $y\in\ \mathbb{R}$, the purpose of symbolic regression is to discover a mathematical expression $f(x_1,x_2,...,x_m)$ through certain methods so that $f$ can fit the data $y$ well. 

In symbolic regression, we usually represent an expression as a binary tree of formulas. Traditional symbolic regression methods are usually based on Genetic programming(GP) methods \cite{haeri2017statistical},\cite{langdon1998genetic},\cite{koza1994genetic}. The main idea of these methods is to randomly initialize a population of expressions, and then imitate the natural evolution process of human beings by crossover and mutation. The disadvantage of these methods is that they are very sensitive to hyperparameters. And because mutation and crossover are completely random, the evolution process is very slow and contains great uncertainty. To alleviate the shortcomings of the GP algorithm, a series of symbolic regression methods based on reinforcement learning (RL)\cite{kaelbling1996reinforcement}, \cite{arulkumaran2017deep}, \cite{li2017deep}, \cite{mnih2015human}  have been proposed, and the most representative of them are DSR and DSO. DSR\cite{petersen2019deep} first randomly initializes a Recurrent Neural Network(RNN) \cite{sherstinsky2020fundamentals}, \cite{yu2019review} as the policy network. The parameters of RNN are then optimized from scratch with the risk policy gradient. DSO is based on DSR by introducing the GP algorithm. The advantages of this kind of method are good versatility, more flexibility, and the ability to adapt to almost any type of data. However, because the RNN has to be trained from scratch for each new formula, the search efficiency of the algorithm is relatively low. Another class of symbolic regression algorithms casts the mapping from data [X, Y] to formula preorder traversal as a translation problem and trains a Transformer model with a large amount of synthetic data. e.g., NeSymReS \cite{biggio2021neural}, SNIP \cite{meidani2023snip}. The advantage of such large-scale pre-training algorithms is that inference speed is relatively fast. However, such models do not generalize well and do not perform well on data outside the training set. Even more, if $X$ is sampled in [-2,2] during training, then when the data is not sampled in this range. For example, [-4,4], or even [-1,1], doesn't work very well. Moreover, the noise robustness performance of such methods is not very good, which leads to the limitation of their application in real life. The above two types of methods have been relatively independent in development. Can we design an algorithm that inherits the advantages of these two approaches while minimizing their disadvantages?


So can we directly use the learning process of a reinforcement learning-based symbolic regression algorithm (DSR, DSO) as training data, and pre-train a Transformer that can directly generate a reinforcement learning training process when given a new [X,Y] input and automatically update the policy in context such that the reward value of the newly obtained expression keeps increasing?

To summarize the above analysis, we are inspired by the Algorithmic Distillation technique (AD)\cite{laskin2022context}, which allows us to train a general reinforcement learning policy on a large amount of offline RL data. We propose a method FormulaGPT, which distills RL algorithms into a transformer by modeling their training histories with a causal sequence model. We use 1.5M learning sequences of DSR and DSO search expressions to train a 'general policy network model' for symbolic regression. So that the model can automatically update the policy in the context, and finally obtain the goal expression.
Specifically, we collect the training process of DSR and DSO, (e.g. [sin, exp, x, and 0.9, cos, exp,..., +, sin, x, x, 1.0]) and sample point [X, Y] as a pair of training data. A data feature encoder SetTransformer is used to extract features from the data [X,Y]. Then the sequence of the learning process is generated in the decoder part. Each time we generate a complete expression, we compute its reward R and then plug R into the sequence to continue generation. For example, now that [sin,exp,x] is a complete expression, we compute the reward R=0.9, and then we add '0.9' to the sequence to get $[sin,exp,x,0.9]$. Then continue generating, and obtain the sequence $[sin, exp, x, 0.9, cos, exp...]$. Until $R^2$ reaches a preset threshold, or the maximum sequence length is reached.

\begin{itemize}
\item We propose a symbolic regression algorithm, FormulaGPT, which not only has good fitting performance on multiple datasets but also has good noise robustness, versatility, and high inference efficiency.
\item We apply the discrete reward sequence of the search process of reinforcement learning-based symbolic regression to train a Transformer, and successfully train a symbolic regression general policy network that can automatically update the policy in context.
\item In addition to using the full history sequence, we also extract a 'shortcut' training data from each training history. That is a path where $R^2$ goes all the way up, with no oscillations. Experiments show that this operation can greatly improve the inference efficiency of formulaGPT.
\item We have greatly improved the shortcomings of previous large-scale pre-training models, such as poor noise robustness and poor versatility so that such models are expected to be truly used to solve practical problems.
\end{itemize}

\section{Relation work}
\subsection{Based on genetic programming}
This kind of method is a classical kind of algorithm in the field of symbolic regression. GP \cite{10.1145/2576768.2598291}, \cite{McConaghy2011}, \cite{7473913} is the main representative of this kind of method, its main idea is to simulate the process of human evolution. Firstly, it initialized an expression population, then generated new individuals by crossover and mutation, and finally generated a new population by fitness. The above process is repeated until the target expression is obtained. RSRM\cite{xureinforcement} integrates the GP algorithm with Double Q-learning\cite{hasselt2010double} and the MCTS algorithm\cite{coulom2006efficient}. a Double Q-learning block, designed for exploitation,  that helps reduce the feasible search space of MCTS via properly understanding the distribution of reward, In short,  the RSRM model consists of a three-step symbolic learning process: RLbased expression search, GP tuning, and MSDB. In this paper\cite{fong2022rethinking}, the fitness function of the traditional GP algorithm is improved, which promotes the use of an adaptability framework in evolutionary SR which uses fitness functions that alternate across generations.
\subsection{Based on reinforcement learning}
Reinforcement learning-based algorithms treat symbolic regression as a combinatorial optimization problem. The typical algorithm is DSR\cite{petersen2019deep}, which uses a recurrent neural network as a policy network to generate a probability distribution P for sampling, and then samples according to the probability P to obtain multiple expressions. The reward value of the sampled expressions is calculated and the policy network is updated with the risky policy, and the loop continues until the target expression is obtained. DSO\cite{mundhenk2021symbolic} is based on DSR by introducing the GP algorithm. The purpose of the policy network is to generate a better initial population for the GP algorithm. Then, the risk policy gradient algorithm is also used to update the policy network. Although the above two algorithms are very good, the efficiency is low, and the expression is more complex, especially the DSO algorithm is more obvious. There have been many recent symbolic regression algorithms based on the Monte Carlo tree search. SPL\cite{sun2022symbolic} uses MCTS in the field of symbolic regression and introduces the concept of modularity to improve search efficiency. However, due to the lack of guidance of MCTS, the search efficiency of this algorithm is low. To improve the search efficiency of the algorithm, the two algorithms DGSR-MCTS \cite{kamienny2023deep} and TPSR \cite{shojaee2024transformer} introduced the policy network to guide the MCTS process based on the previous algorithm. While maintaining the performance of the algorithm, it greatly improves the search efficiency of the algorithm. However, although the above two algorithms improve the search efficiency of the algorithm, they reduce the Versatility of the algorithm, and the noise robustness ability of the algorithm is also greatly reduced. To solve the above problems and balance the Versatility and efficiency of the algorithm, SR-GPT \cite{li2024discovering} uses a policy network that learns in real-time to guide the MCTS process. It achieves high performance while efficient search.
\begin{figure*}[tp]
\centering
\includegraphics[width=148mm]{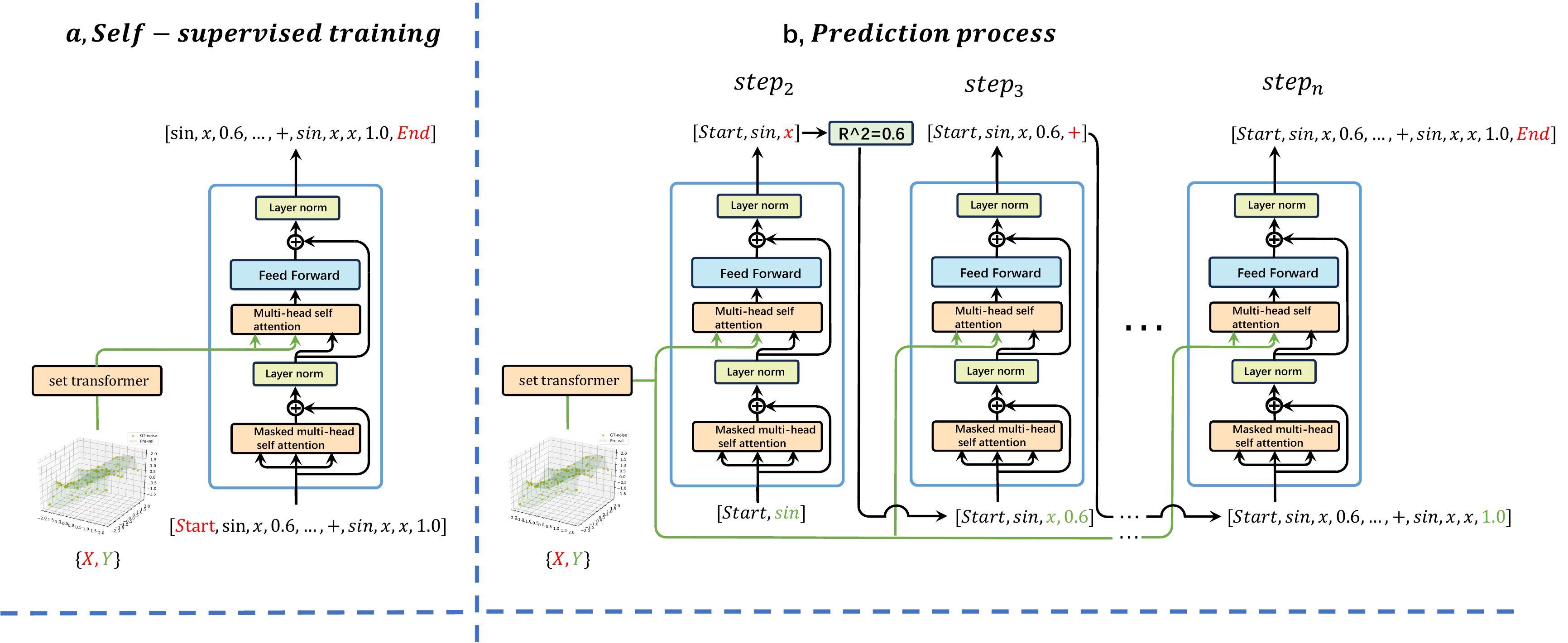}
\caption{
Figure (a) shows the training process of FormulaGPT; Figure (b) shows the inference process of FormulaGPT.
}
\label{fig1}
\end{figure*}
\subsection{Based on pre-training}
Many SR methods based on reinforcement learning have good Versatility. However, its search efficiency is relatively low, and it often takes a long time to get a good expression. In contrast, pre-trained models treat the SR problem as a translation problem and train a transformer with a large amount of artificially synthesized data in advance. Each prediction only needs one forward propagation to get the result, which is relatively efficient. SymbolicGPT\cite{valipour2021symbolicgpt} was the first large-scale pre-trained model to treat each letter in a sequence of symbols as a token, (e.g.['s',' i','n', '(', 'x', ')']). A data feature extractor is used as the encoder, and then each token is generated by the Decoder in turn. Finally, the predicted sequence and the real sequence are used for cross-entropy loss. BFGS is used to optimize the constant at placeholder 'C'. NeSymReS\cite{biggio2021neural} builds on symbolicGPT by not thinking of each individual letter in the sequence of expressions as a token. Instead, Nesymres represents the expression in the form of a binary tree, which is then expanded by preorder traversal, and considers each operator as a token (e.g. ['sin','x']). Then SetTransformer is used as the Encoder of the data, and finally, Decoder is used to generate the expression sequence. The overall framework and idea of the EndtoEnd\cite{kamienny2022end} algorithm are not much different from NeSymReS, but EndtoEnd abandons the constant placeholder 'C', encodes the constant, and directly generates the constant from the decoder. The constants are then further optimized by Broyden-Fletcher-Goldfarb-Shanno (BFGS) \cite{liu1989limited}. 
Symformer\cite{vastl2024symformer} is slightly different from the previous pre-trained models in that it directly generates the constant values in the expression as well as the sequence of expressions. SNIP\cite{meidani2023snip} first applies contrastive learning to train the feature encoder and then freezes the encoder to train the decoder. But SNIP works well only when combined with a latent space optimization (LSO)\cite{bojanowski2017optimizing} algorithm. MMSR\cite{li2024mmsr} solves the symbolic regression problem as a pure multimodal problem, takes the input data and the expression sequence as two modalities, introduces contrastive learning in the training process, and adopts a one-step training strategy to train contrastive learning with other losses.
\subsection{Based on deep learning}
This class of methods combines symbolic regression problems with artificial neural networks, where EQL replaces the activation function in ordinary neural networks with [sin, cos,...] And then applies pruning methods to remove redundant connections and extract an expression from the network. EQL\cite{kim2020integration} is very powerful, however, it can't introduce division operations, which can lead to vanishing or exploding gradients.
The main idea of AI Feynman 1.0 \cite{udrescu2020ai} and AI Feynman 2.0\cite{udrescu2020ai} series algorithms are to “Break down the complex into the simple”. by first fitting the data with a neural network, and then using the trained neural network to discover some properties (e.g. Symmetry, translation invariance, etc.) to decompose the function hierarchically. AI Feynman 2.0 introduces more properties based on AI Feynman 1.0, which makes the scope of its application more extensive relative to AI Feynman 1.0. MetaSymNet\cite{li2023metasymnet} takes advantage of the differences between symbolic regression and traditional combinatorial optimization problems and uses more efficient numerical optimization to solve symbolic regression. 
\begin{table*}[htp]
\renewcommand{\arraystretch}{1.1}
\vspace{-0.2cm}
\centering
\caption{At a 0.95 confidence level, a comparison of the coefficient of determination ($R^2$) was conducted between FormulaGPT and four baseline models. Bold values signify state-of-the-art (SOTA) performance.
\label{tab1}}
\resizebox{14.2cm}{!}{
\begin{tabular}{ccccccc}
\toprule[1.45pt]
\toprule
Dataset & Name  & FormulaGPT & DSO & SNIP & SPL& NeSymReS  \\ 
\toprule
Dataset-1 & Nguyen    & $\textbf{0.9999}_{\pm0.001}$ & $\textbf{0.9999}_{\pm0.001}$ & $0.9936_{\pm0.003}$& $0.9842_{\pm0.001}$& $0.8468_{\pm0.002}$  \\
Dataset-2 & Keijzer   & $\textbf{0.9962}_{\pm0.003}$ & $0.9924_{\pm0.001}$& $0.9862_{\pm0.002}$ & $0.8919_{\pm0.002}$ & $0.7992_{\pm0.002}$  \\
Dataset-3 & Korns     & $0.9801_{\pm0.003}$  &$ \textbf{0.9872}_{\pm0.002}$& $0.9418_{\pm0.004}$  & $0.8788_{\pm0.001}$ & $0.8011_{\pm0.001}$  \\
Dataset-4 & Constant  & $\textbf{0.9981}_{\pm0.002}$  & $0.9928_{\pm0.003}$ & $0.9299_{\pm0.002}$  & $0.8942_{\pm0.002}$  & $0.8444_{\pm0.002}$   \\
Dataset-5 & Livermore  & $\textbf{0.9815}_{\pm0.003}$ & $0.9746_{\pm0.002}$& $0.8948_{\pm0.004}$& $0.8728_{\pm0.002}$ & $0.7136_{\pm0.0004}$ \\
Dataset-6 & Vladislavleva  & $0.9782_{\pm0.002}$ & $\textbf{0.9963}_{\pm0.004}$ & $0.9212_{\pm0.005}$ & $0.8433_{\pm0.004}$ & $0.6892_{\pm0.004}$  \\
Dataset-7 & R  & $\textbf{0.9828}_{\pm0.004}$ & $0.9744_{\pm0.004}$ & $0.9614_{\pm0.001}$ & $0.9122_{\pm0.003}$ & $0.8003_{\pm0.004}$  \\
Dataset-8 & Jin  & $\textbf{0.9912}_{\pm0.002}$ & $0.9823_{\pm0.003}$ & $0.9877_{\pm0.004}$ & $0.9211_{\pm0.002}$& $0.8627_{\pm0.002}$\\
Dataset-9 & Neat  & $\textbf{0.9962}_{\pm0.002}$& $ 0.9827_{\pm0.005}$ & $ 0.9401_{\pm0.004}$ & $ 0.8828_{\pm0.003}$& $ 0.7996_{\pm0.005}$ \\
Dataset-10 & Others  & $\textbf{0.9936}_{\pm0.003}$ & $0.9861_{\pm0.003}$& $0.9702_{\pm0.003}$ & $0.9435_{\pm0.003}$ & $0.8226_{\pm0.002}$ \\
\toprule
SRBench-1 & Feynman  & $\textbf{0.9940}_{\pm0.002}$ &$0.9610_{\pm0.004}$ &$0.8899_{\pm0.003}$  & $0.9284_{\pm0.003}$ &$0.7025_{\pm0.003}$\\
SRBench-2 & Strogatz  & $\textbf{0.9807}_{\pm0.004}$ &$0.9313_{\pm0.003}$ &$0.8307_{\pm0.003}$  & $0.8411_{\pm0.002}$ &$0.6222_{\pm0.002}$\\
SRBench-3 & Black-box & $0.9601_{\pm0.003}$ &$\textbf{0.9833}_{\pm0.002}$ &$0.8692_{\pm0.003}$  & $0.9024_{\pm0.002}$ &$0.6825_{\pm0.003}$\\
\cline{2-7} 
 & \multicolumn{1}{r}{Average} & $\textbf{0.9870}$ &$0.9803$ &$0.9321$  & $0.8997$ & $0.7682$
\end{tabular}
}
\end{table*}

\section{Method}
We train our FormulaGPT with the 1.5M data collected from the DSR and DSO training processes. FormulaGPT uses a SetTransformer\cite{lee2019set} to extract the features of the feature data, and then Decoder generates the sequence of the DSR training process.  After training with massive data, our model distills reinforcement learning into a Transformer. Then we only need to provide data, and FormulaGPT can quickly generate the training process that takes a long time for DSR (DSO) in an autoregressive way. The overall structure of our algorithm is shown in figure \ref{fig1}.
\subsection{Expressions generation}
In FormulaGPT, we use symbols [$+, -, *, /, sin, cos, log, sqrt, C, x_1, x_2,..., x_n$]. Where C denotes a constant placeholder (e.g. sin(2.6*x) can be written as sin(C*x), and preorder traversal is [sin, *, C, x]). And $[x_1,...,x_n]$ denotes a variable. Expressions composed of the above symbols can be expressed in the form of a binary tree, and then according to the preorder traversal expansion of the binary tree, we can obtain a sequence of expressions.
\subsubsection{Arity(s) function}
We use these symbols to randomly generate sequences of expressions. Specifically, we introduce the Arity(s) function if s is a binary operator. e.g. $[+, -, *, /]$, then Arity(s)=2; Similarly, if s is a unary operator. e.g. $[sin, cos, exp, log, sqrt]$, then Arity(s)=1; If s is a variable $[x_1,...,x_n]$ or a constant placeholder `C', then Arity(s)=0. 
\subsubsection{Generation stop decision}
Before we start generating the expression, we import a counter `count' and initialize it to 1. Then we randomly take a symbol s from the symbol store and update the count value according to the formula $count= count-Arity(s)-1$ until count=0. At this point, we have a complete sequence of expressions.
\subsubsection{Generation constraints}
In the process of expression generation, we need to ensure that the generated expression is meaningful. Therefore, we make the following restriction. (1), Trigonometric functions cannot be nested.(e.g. sin(cos(x)) ). because in real life, this form rarely occurs.
(2), For functions like log(x) and sqrt(x), you can't have a negative value at x. e.g.log(sin(x)),sqrt(cos(x)), because both sin(x) and cos(x) can take negative values.
\begin{figure*} [htp]
    \centering
    \setlength{\belowcaptionskip}{-0.2cm} 
	  \subfloat[]{       
        \includegraphics[width=0.48\linewidth]{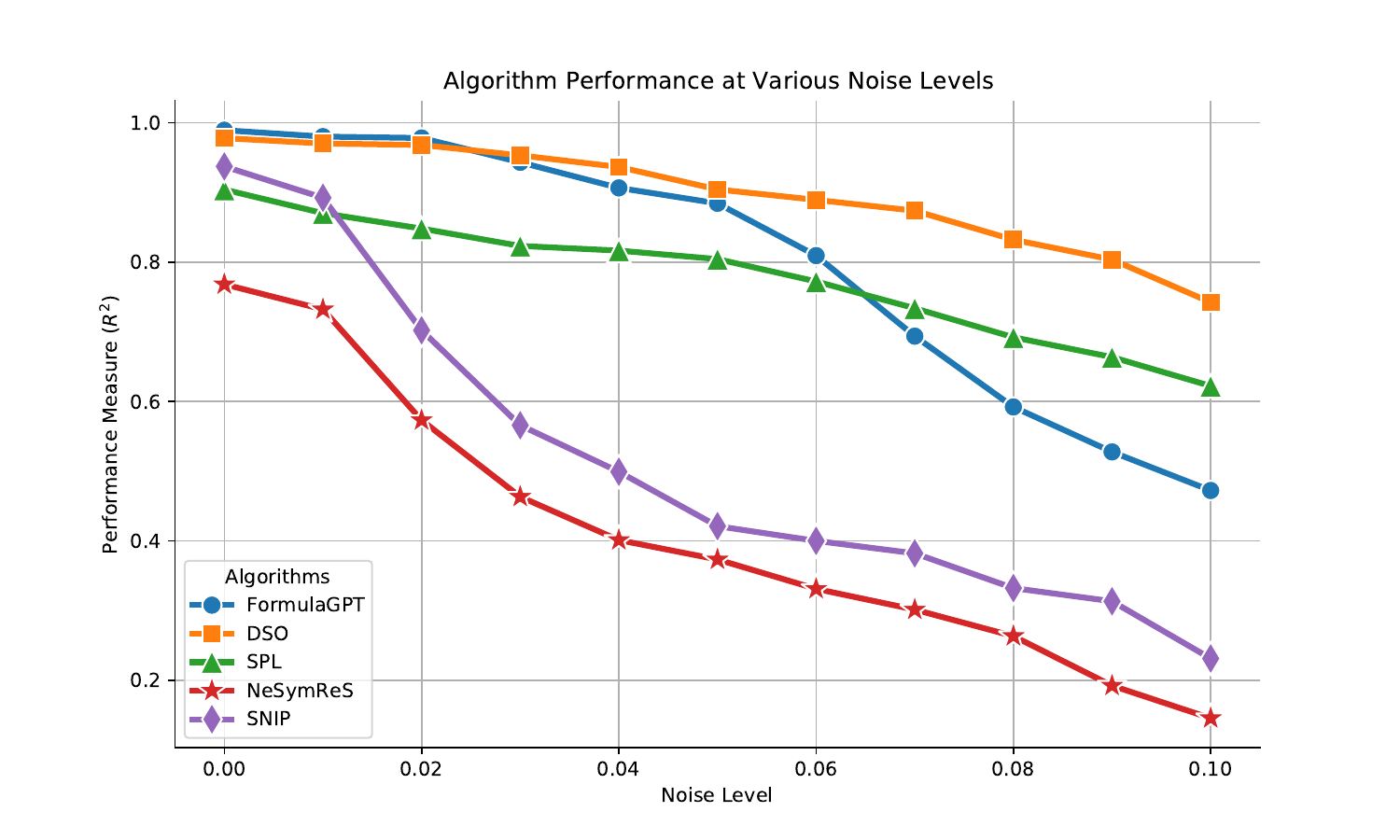}\label{fig2a}}
        \subfloat[]{
        \includegraphics[width=0.44\linewidth]{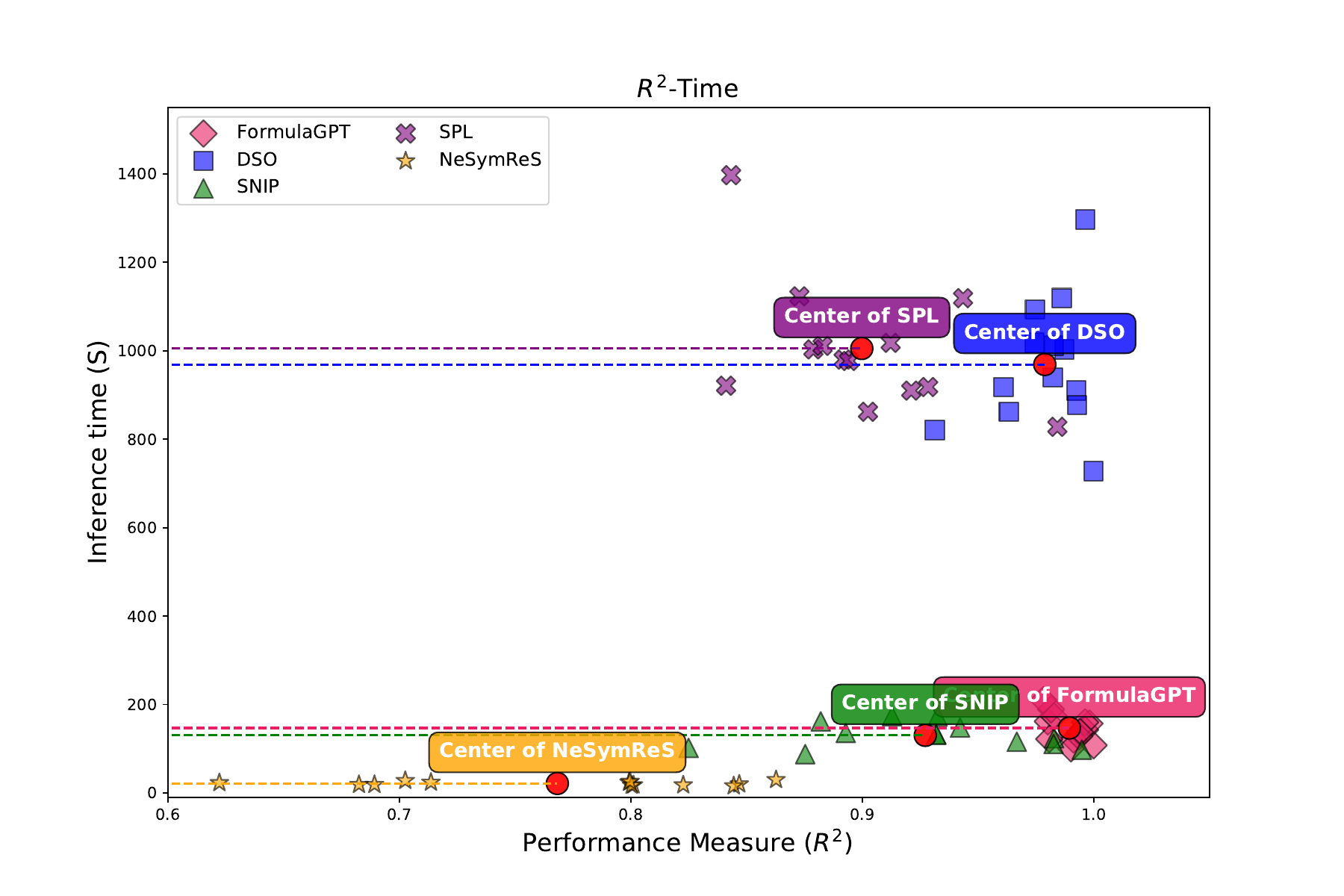}\label{fig2b}}   
	  \caption{
   Figure (a) shows the noise robustness ability of the five methods. From the figure, we can see that although the noise robustness of FormulaGPT is not as good as that of DSO, it is better than that of SNIP and NeSymReS. Figure (b) shows the $R^2$-time scatter plot. From the figure, we can see that the inference time of FormulaGPT is much lower than that of DSO and SPL, and slightly slower than that of SNIP and NeSymReS. In particular, the closer the center point of each algorithm is to the bottom right corner, the better the comprehensive performance of the algorithm is. FormulaGPT is one of the algorithms closest to the bottom right.}
\label{fig2} 
\end{figure*}
\begin{figure*}[htp]
\centering
\includegraphics[width=145mm]{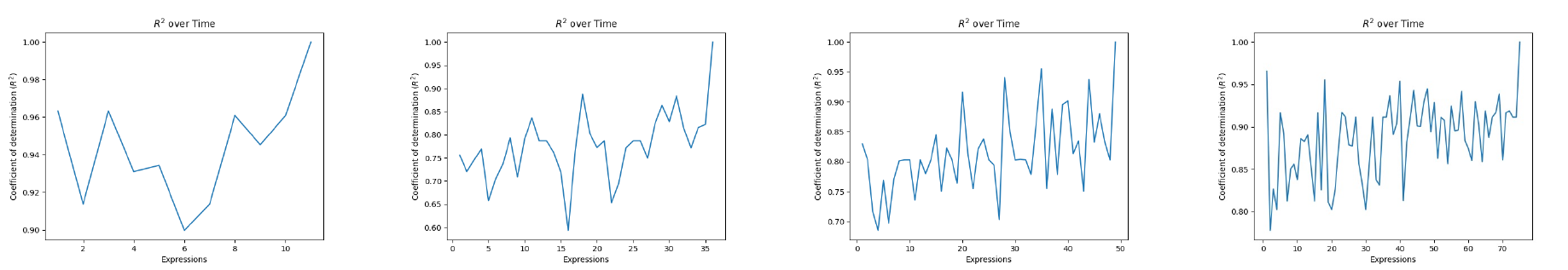}
\caption{
This figure shows the change of $R^2$ over time when FormulaGPT searches some expressions. From the figure, we can see that although $R^2$ has oscillation, the overall $R^2$ still shows an upward trend.
}
\label{fig3}
\end{figure*}
\subsection{Training data collection}
After obtaining the skeleton of the expression, we sample X in the interval [-10,10] and compute the corresponding y.
These sampled data [X,y] are then fed into DSR and DSO to search for results. During the search process, we collect the search process sequence of DSR and DSO (e.g. $[sin, x_1, 0.63, cos, x_1, 0.42, +, sin, x_1, cos, x_2, 0.84,..., +, sin, x_1, x_2, 1.0]$). Note that DSR and DSO sample multiple expressions at a time, and we only select the expression with the largest $R^2$\cite{ozer1985correlation} as the sampled expression, and join the collection sequence with it and its $R^2$. When there is an expression $R^2$ greater than 0.99, we stop the search and save the collected process sequence and data in pairs. The maximum token length of our process sequence is 1024, and if $R^2$ does not exceed 0.99 beyond this length, we terminate the search and discard the pair of data. For $R^2$ we set up a total of 100 numerical [0.00, 0.01, 0.02,..., 0.98.0.99, 1.00]. Round the $R^2$ of each expression to two decimal places. $R^2$ less than zero is set to 0.00. Then it is converted to a string type and spliced into the collection sequence. In particular, to improve the inference efficiency of the algorithm and make FormulaGPT have the ability to surpass DSO, we extract a `shortcut' from each training data collected above, so that the original oscillating rising $R^2$ becomes rising all the time. For example, for the above example, we would remove the oscillating $R^2$ of 0.42 and the sequence would be $[sin, x_1, 0.63, +, sin, x_1, cos, x_2, 0.84,..., +, sin, x_1, x_2, 1.0]$. $R^2$ is going to keep going up. See the Appendix\ref{number-of-expressions} for ablation experiments and proof of effectiveness on the 'Shortcut' dataset.

\subsection{Model architecture}
\subsubsection{Data encoder}
The data information plays an important role in guiding the Decoder. To accommodate the data feature's permutation invariance, wherein the dataset's features should remain unchanged regardless of the order of the input data, we utilize the SetTransformer as our data encoding method, as described by \cite{pmlr-v97-lee19d}. Our encoder takes a set of data points $\mathcal{D}=\{X,y\}\in\mathbb{R}^{n\times d}$. These data points undergo an initial transformation via a trainable affine layer, which uplifts them into a latent space $h_n\in\mathbb{R}^{d_h}$. Subsequently, the data is processed through a series of Induced Set Attention Blocks (ISABs)\cite{pmlr-v97-lee19d}, which employ several layers of cross-attention mechanisms. Initially, a set of learnable vectors serves as queries, with the input data acting as the keys and values for the first cross-attention layer. The outputs from this first layer are then repurposed as keys and values for a subsequent cross-attention process, with the original dataset vectors as queries. Following these layers of cross-attention, we introduce a dropout layer to prevent overfitting. Finally, the output size is standardized through a final cross-attention operation that uses another set of learnable vectors as queries, ensuring that the output size remains consistent and does not vary with the number of inputs.
\subsubsection{Training sequence decoder}
Training sequence decoder We adopt the standard transformer\cite{vaswani2017attention} decoder architecture. The training sequence first passes through the multi-head attention mechanism, and the Mask operation is used here to prevent later information leakage. Then another multi-head attention mechanism introduces the cross-attention mechanism to interact with the data features extracted by SetTransformer and the Training sequence features. 

\begin{figure*}[h]
\centering
\hspace{-2.6em}
\includegraphics[width=148mm]{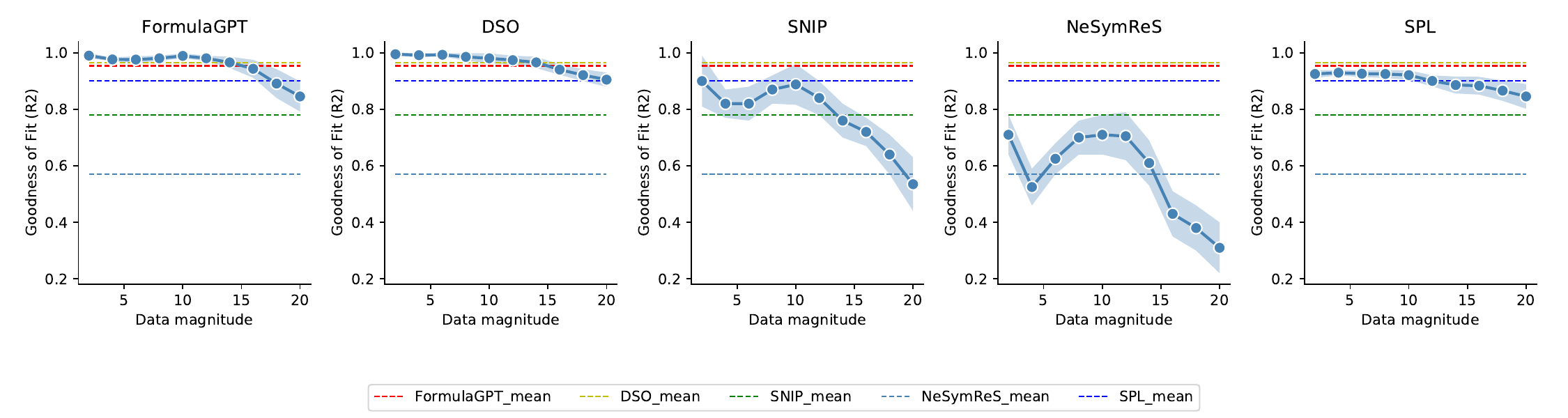}
\caption{
This figure shows the Versatility test of five algorithms. From the variation trend of $R^2$ in the figure, we can see that although FormulaGPT is not as general as DSO and SPL, it is far better than SNIP and NeSymReS two pre-training methods. We achieved what we expected.
}
\label{fig4}
\end{figure*}
\subsection{Model training}
We train FoemulaGPT on the ability to generate a target expression by automatically adapting its policy in context as prompted by the data $\mathcal{D}=\{X,y\}\in\mathbb{R}^{n\times d}$. Specifically, formulaGPT mainly contains an Encoder and Decoder. Where the Encoder is SetTransformer, we take $\mathcal{D}=\{X,y\}\in\mathbb{R}^{n\times d}$ input, get a feature $z$, and then use this feature $z$ as a hint to guide the Decoder to generate the corresponding DSR training process sequence. For example, We have data [X,y], and a corresponding DSR training sequence $S=[sin,x,0.32,+, sin, x, cos, x, 0.8,...,+, sin, x, x, 1.0]$. [X,y] is fed into SetTransformer to get feature $z$. The decoder generates a sequence $\hat{S}$ given $z$: it produces a probability distribution $P(\hat{S}_{k+1}|\hat{S}_{1:k},z)$ over each token, given the previous tokens and $z$. Finally, the cross-entropy loss\cite{zhang2018generalized}, \cite{mao2023cross} between S and $\hat{S}$ is calculated.

After the training of massive data, the symbolic regression method based on reinforcement learning is distilled into a transformer. Once trained, and given new data, the transformer can automatically update the policy in context, the $R^2$ of the generated expression shows an overall upward trend until the stopping condition is reached. Enables fast completion of the reinforcement learning training process in context.

\subsection{Constant optimization}
During decoder generation, every time a complete expression is generated, we will compute its $R^2$, which will involve the constant optimization problem. For example, if we get a preorder traversal of an expression, [*, C, sin, x], the corresponding expression is C*sin(x), then we need to use the BFGS algorithm to optimize the constant value at C with X as input and y as output.
\subsection{Prediction process}
After FormulaGPT is trained, with a new set of data [X,y], we can automatically simulate the search reinforcement process of reinforcement learning in the context, and then obtain the target expression. Specifically, the data [X,y] is fed into the SetTransformer to get the feature z, and then the Decoder gets  $P(\hat{S}_{1}|start,z)$ to generate the first symbol. 
Then, $P(\hat{S}_{k+1}|\hat{S}_{1:k},z)$ is used to generate new symbols in turn until a complete expression is obtained, then $R^2$ is calculated, and $R^2$ is incorporated into the existing sequence, and new symbols are generated in the same way until $R^2>0.99$ or the sequence reaches its maximum length. A schematic diagram of the prediction process is shown in Figure \ref{fig1}b.

\section{Experiment}
To verify whether FormulaGPT achieves what we expect: while inheriting the advantages of the SR algorithm based on reinforcement learning (e.g Strong noise robustness and Versatility) and the advantages of the pre-training-based SR algorithm (e.g High inference efficiency), and can maintain good comprehensive performance. To make the experiment more convincing, we collected more than a dozen SR datasets including SRBench as the test set. The dataset mainly includes: 'Nguyen', 'Keijzer', 'Korns', 'Constant', 'Livermore', 'Vladislavleva', 'R',' Jin', 'Neat',' Others', 'Feynman',  'Strogatz' and 'Black-box'. The three datasets Feynman ', 'Strogatz' and 'Black-box' are from the SRbetch dataset.

We compare SR-GPT with four symbol regression algorithms that have demonstrated high performance:
\begin{itemize}
\item \textbf{DSO}. A symbolic regression algorithm that deeply integrates DSR with the GP algorithm
\item \textbf{SPL}. An excellent algorithm that successfully applies the traditional MCTS to the field of symbolic regression.

\item \textbf{NeSymReS}. This algorithm is categorized as a large-scale pre-training model.
\item \textbf{SNIP}. A large-scale pre-trained model with a feature extractor trained with contrastive learning before training. 
\end{itemize}

\subsection{Comparison of $R^2$}
The most important goal of symbolic regression is to find an expression from the observed data that accurately fits the given data. A very important indicator to judge the goodness of fit is the coefficient of determination ($R^2$).  Therefore, we tested the five algorithms on more than ten datasets, using $R^2$ as the standard. We run each expression in the dataset 20 times and then take the average of all the expressions in the dataset. And the confidence level\cite{junk1999confidence}, \cite{costermans1992confidence} is taken to be 0.95. The specific results are shown in Table \ref{tab1}. As we can see from the table, FormulaGPT is not optimal except on individual datasets. However, on the whole, FormulaGPT performs slightly better than the other four baselines. From the table, we can see that FormulaGPT performs worse than DSO on two datasets, Korns and Vladislavleva. According to our analysis, the expressions in these two data sets are relatively complex, and FormulaGPT may not perform well in overly complex expressions due to the limited training data. 

\subsection{Comparison of noise robustness performance}
One of the biggest advantages of DSO and SPL over pre-trained algorithms is their strong robustness to noise. The reason is that for each new observation (noisy or not), DSO and SPL have to train from scratch and eventually find an expression that fits the data well. 
The goal of FormulaGPT is to combine the advantages of the above two types of algorithms as much as possible, that is, the high efficiency of the pre-training algorithm and the noise robustness of the reinforcement learning algorithm. 

Therefore, to test the noise robustness of FormulaGPT, we tested the noise robustness of five algorithms on the datasets and made a comparison. Specifically, we add ten different levels of noise to each data of the Nguyen dataset. $y_{noise} = y + L*|(max(y)-min(y)| * D_{noise} $. Where $L$is the noise level value [0.00, 0.01, 0.02,..., 0.10], $|(max(y)-min(y)|$ is the span of the current data. These two values can control the magnitude of the added noise. $D_{noise}$ is Gaussian noise normalized to the interval [-1,1]. The test results of the five algorithms are shown in Fig.\ref{fig2a}. From the figure, we can see that even though our algorithm is slightly weaker than DSO in noise robustness performance, it is basically on par with SPL and far better than SNIP and NeSymReS, two pre-training algorithms.
\subsection{Comparison of Versatility}
Real-world data can range from [-1,1] to [-5,10]. However, the pre-trained large model needs to specify a sampling range when sampling the training data, and once it goes beyond this range, its performance will drop significantly. For example, if all of our training data is sampled between [-10,10], then when we test with data between [-15,15], the model will not perform well. Even if we take values between [-5,5], the model doesn't work very well. In theory, the SR algorithm based on reinforcement learning does not have this problem, because it learns from scratch for each piece of data.
formulaGPT attempts to distill reinforcement learning algorithms into Transformers utilizing large-scale pre-training. Then, given a new set of data, formulaGPT generates a reinforcement learning process and automatically updates the policy in context. Therefore, FormulaGPT should inherit the advantages of good Versatility of the SR algorithm based on reinforcement learning. To test the Versatility of FormulaGPT. All of our training data is sampled in the interval [-10,10], (SNIP and NeSymReS as well). Then in [-2,2],[-4,4],...,[-20,20] The ten intervals are sampled and tested separately. The results of the Versatility test of the five algorithms are shown in figure \ref{fig4}.

From figure \ref{fig4}, we can find that DSO and SPL are not too sensitive to the interval, but their performance will also be affected when the interval becomes larger. However, the pre-trained algorithms SNIP and NeSymReS only perform well in the interval [-10,10], and their performance is relatively worse as the interval is further away from 10. Especially if it's bigger than 10. FormulaGPT has little influence when the interval is less than 10, and when the interval is greater than 10, although it also has influence, the extent is far less than that of SNIP and NeSymReS. This shows that our algorithm meets our expectations and makes the ability of the pre-trained model greatly improved in terms of Versatility.

It is worth noting that both pre-training methods perform well in the interval [-1,1]. This is because the interval [-1,1] is too small and many of the expression curves are simple and similar.
\subsection{Comparison of inference time}
Inference time is an important indicator for evaluating symbolic regression methods. To test the search efficiency of the algorithms, we plotted the $R^2$-time scatter plots of each algorithm for all the datasets. For each algorithm, the termination search condition we set is $R^2>0.99$ or reach the termination condition of the algorithm itself (FormulaGPT: reach the maximum length; DSO: up to 400 epochs; SPL: MCTS process is executed 400 times; SNIP: performs inference once NeSymReS: Performs inference once). From the figure, we can find that FormulaGPT's inference speed is not as fast as NeSymReS and SNIP, but it is much faster than DSO and SPL. This also meets our expectations. In particular, in Figure \ref{fig2b}, for each algorithm, the closer its center point is to the bottom right of the picture, the better its overall performance is. We can find that formulaGPT is one of the algorithms closest to the bottom right corner.
\subsection{The $R^2$ trend of the generation process}
We train a transformer using a large number of search histories of SR algorithms based on reinforcement learning as training data and try to distill reinforcement learning into the transformer. That is, for a new set of data, the Decoder has already completed the policy update in the context during the sequence generation process. Furthermore, the $R^2$ of the formula generated in the process should show an upward trend of shock as a whole, just like reinforcement learning. To prove whether formulaGPT meets our expectations, we plot the trend of $R^2$ of the expressions obtained by FormulaGPT during the search for Nguyen-1, Nguyen-2, Nguyen-3, and Nguyen-4. It can be seen from Figure \ref{fig3} that in the process of expression generation, although $R^2$ fluctuates somewhat, its overall trend is still upward. Our intended goal of automatically updating the policy in context is achieved.
\section{Discussion and Conclusion}
In this paper, we propose FormulaGPT, a new symbolic regression method. We collected many learning histories of DSR search target expressions as training data and then trained a Transformer with these data. Our goal is to distill reinforcement learning-based SR algorithms into Transformer so that Transformer can directly generate a reinforcement learning search process and automatically update policies in context. And even reach a new level of performance. 

By testing on more than a dozen datasets including SRBench, FormulaGPT achieves state-of-the-art results on several datasets. More importantly, FormulaGPT achieves a good balance between noise robustness Versatility, and inference efficiency while achieving good fitting performance. This makes FormulaGPT have the advantages of both SR algorithms based on large-scale pre-training and reinforcement learning algorithms. This is important because previous large-scale pre-training methods work well on noiseless data, but the data in the real world is almost always noisy, so currently pre-trained models are almost impossible to apply in real life. However, FormulaGPT is expected to break this limitation.
Because it can ensure fast inference while maintaining good noise robustness performance. 

Although FormulaGPT achieves good performance, in the noise robustness experiment, we can find that compared with DSO and SPL, it has a large room for improvement. Next, we will take a deeper look at the data feature extractor to enhance the noise robustness ability of the data. 

\bibliographystyle{plainnat}  
\bibliography{neurips_2024.bib}
\medskip

\newpage
\appendix
\onecolumn
\section{Detailed Settings of hyperparameters during training and inference}
\subsection{Detailed Settings of the hyperparameters of SetTransformer}
\begin{table*}[htbp]
\centering
{
\begin{tabular}{lc}
\textbf{hyperparameters} & \textbf{Numerical value}\\ 
\toprule
 \textbf{N\_p}  & 0\\
 \textbf{activation}  & 'relu'\\
 \textbf{bit16}  & True\\
 \textbf{dec\_layers}  & 5\\
 \textbf{dec\_pf\_dim}  & 512\\
 \textbf{dim\_hidden}  & 512\\
 \textbf{dim\_input}  & 3\\
 \textbf{dropout}  & 0\\
 \textbf{input\_normalization}  & False\\
 \textbf{length\_eq}  & 60\\
 \textbf{linear}  & False\\
 \textbf{ln}  & True\\
 \textbf{lr}  & 0.0001\\
 \textbf{mean}  & 0.5\\
 \textbf{n\_l\_enc}  & 5\\
 \textbf{norm}  & True\\
 \textbf{num\_features}  & 20\\
 \textbf{num\_heads}  & 8\\
 \textbf{num\_inds}  & 50\\
 \textbf{output\_dim}  & 60\\
 \textbf{sinuisodal\_embeddings} & False\\
 \textbf{src\_pad\_idx} & 0\\
 \textbf{std} & 0.5\\
 \textbf{trg\_pad\_idx} & 0\\
 \hline
 \end{tabular} 
\caption{Hyperparameters of SetTransformer}
\label{a-tab6}
}
\end{table*}

\subsection{Detailed Settings of the hyperparameters of the decoder of Transformer}

\begin{table*}[htbp]
\centering
{
\begin{tabular}{lc}
\textbf{hyperparameters} & \textbf{Numerical value}\\ 
\toprule
\textbf{Batchsize} & 128 \\
\textbf{epochs } &  1000\\
\textbf{Embedding Size}  & 512\\
\textbf{Dimension of K( =Q), V}  & 512\\
\textbf{Number of Encoder of Decoder Layer}  & 16\\
\textbf{Number of heads in Multi-Head Attention}  & 16\\
\textbf{Maximum sequence length} & 2048\\
\textbf{momentum}  &0.99\\
\hline
\end{tabular} 
\caption{Hyperparameters of the Decoder of Transformer.}
\label{a-tab6}
}
\end{table*}
\subsection{Hyperparameters of the inference process}

\begin{table*}[htbp]
\centering
{
\begin{tabular}{lc}
\textbf{hyperparameters} & \textbf{Numerical value}\\ 
\toprule
\textbf{Maximum Sequence expression length} & Define yourself \\
\textbf{Maximum sequence length} & 2048\\
\textbf{beam\_width } & 16\\
\hline
\end{tabular} 
\caption{Hyperparameters of the Decoder of Transformer.}
\label{a-tab6}
}
\end{table*}

\newpage
\section{Pseudo-code of the training process for FormlulaGPT.}


\begin{algorithm}[htb]
   \caption{FormulaGPT pre-training}
   \label{alg1}
   \begin{algorithmic}[1] 
   \State \textbf{Input:} data: $[X,y]$; Expression Skeleton: $S$.
   \Repeat
   \State Initialize: FormulaGPT = $ GPT_\theta$.
   \For{$i=1$ \textbf{to} $E$} 
   \For{$j=1$ \textbf{to} $N$} 
   \State $[X_j, y_j]$, $[S_j]$ = GetBatch($[X,y]$, $S$)
   \State $F_{Data}$ = SetTransformer($[X_j, y_j]$)
   \State $F_{Skeleton}$ = SkeletonEncoder($S_j$)
   \State $S_{pred}$ = ExpressionDecoder($F_{Data}$, $F_{Skeleton}$)
   \State $\mathcal{L}_{CE}$ = CrossEntropy($S_j$, $S_{pred}$)
   \EndFor 
   \State Compute the gradient $\nabla_{\theta} \mathcal{L}_{CE}$ and use it to update $\theta$.
   \EndFor 
   \Until{the termination condition is met}
   \end{algorithmic}
\end{algorithm}

\section{Appendix: Effect of training data size on performance}
Figure \ref{figA1} below shows the effect of the training data size on the performance of the algorithm. We take five different sizes of training data of 10k,50k,500k,0.5M, and 1.5M respectively to train FormulaGPT and test on the data sets. From the following figure, we can clearly see that with the increase in data size, the performance of FormulaGPT also gradually becomes better. This is in line with our expectations and also proves that FormulaGPT can achieve even more amazing results if we have a larger scale of training data.
\begin{figure*}[htp]
    \centering
       \subfloat[]{
       \includegraphics[width=0.99\linewidth]{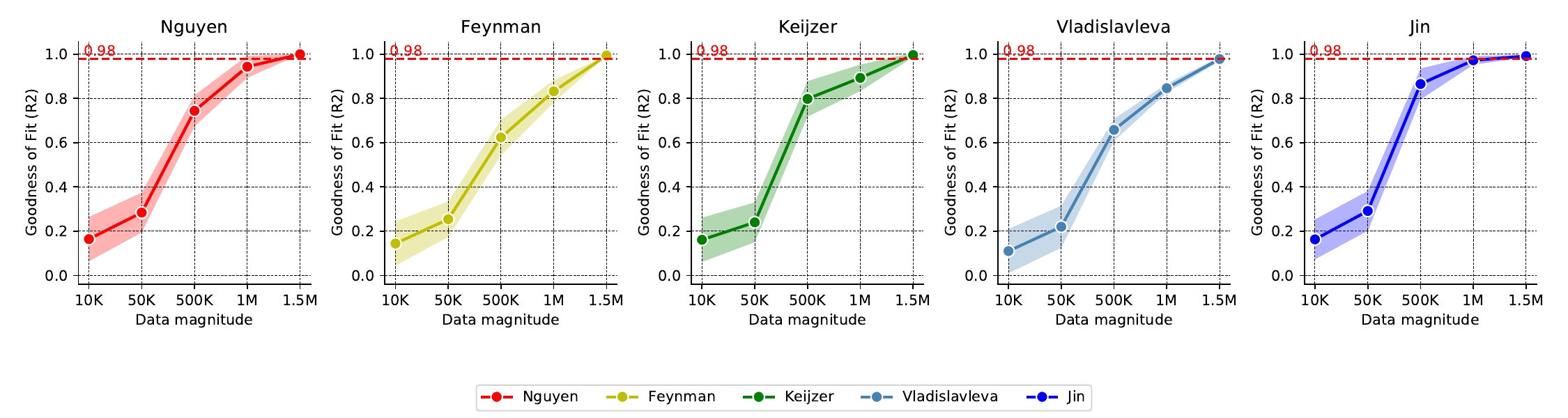} \label{figA1a}}\\
        \subfloat[]{
        \includegraphics[width=0.99\linewidth]{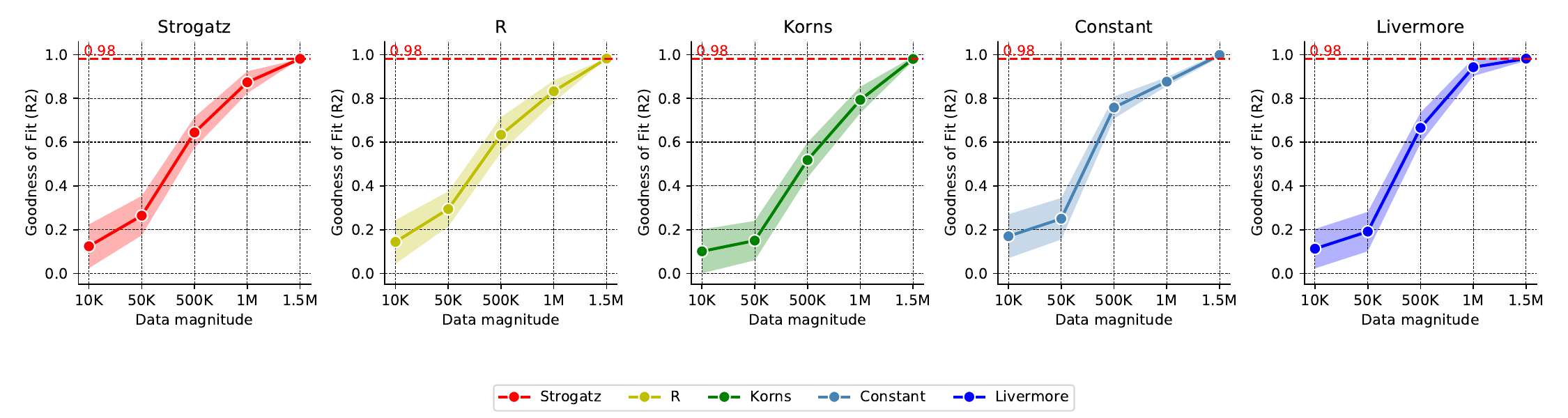}\label{figA2b}}\\
        \subfloat[]{
       \includegraphics[width=0.80\linewidth]{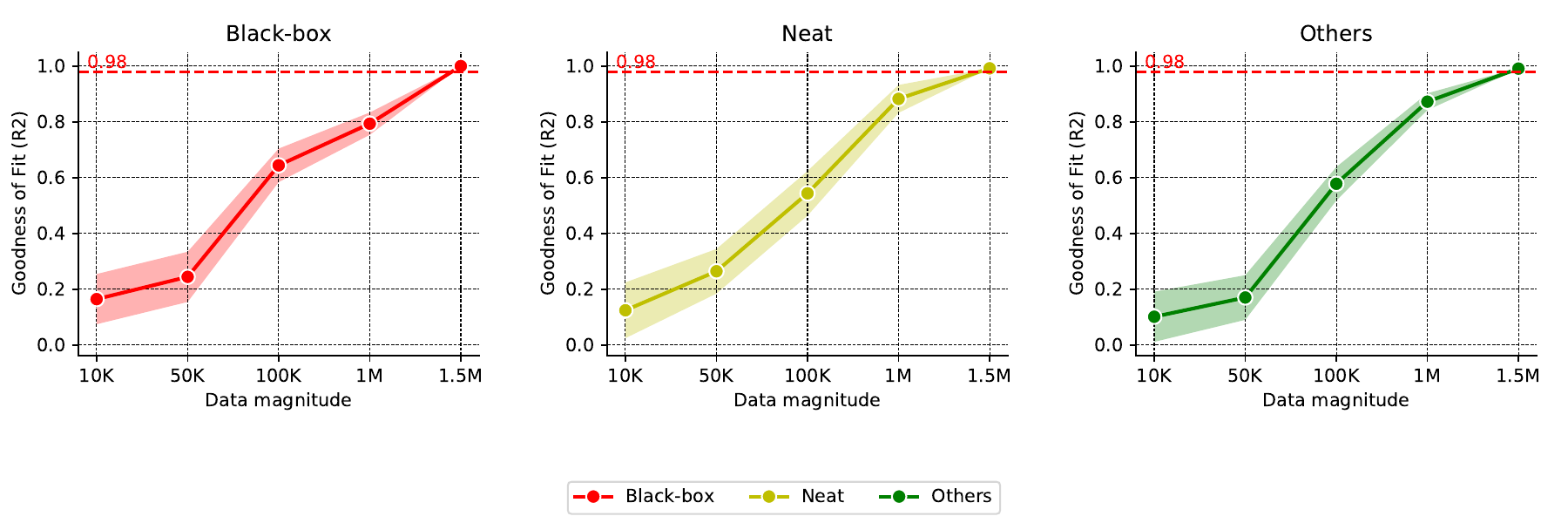} \label{figA3c}}
	  \caption{
   From the figures above, we can see that the performance of the algorithm does not improve much when increasing the data size from 10k to 50k. However, when the data size is from 50k to 500k, the performance of the algorithm has been greatly improved, because the data size is already considerable. Then, with the further increase of data size, the performance of the algorithm also shows a steady upward trend, but the upward speed slows down. }
\label{figA1} 
\end{figure*}
\section{Appendix: Ablation experiments on the `Shortcut' dataset and proof of effect}
\label{number-of-expressions}
The number of expressions in the inference process refers to the number of expressions experienced by the algorithm from the initial expression to the final result expression during the inference process. For example, for the expression cos(x)+x, the inference is sin(x) $-->$ cos(x)$-->$ cos(x)+x. We count the number of its process expressions as three. The index can reflect the reasoning accuracy of the algorithm, and can also reflect the reasoning efficiency of the algorithm from the side. The smaller the number of expressions experienced in the inference process, the higher the inference accuracy and efficiency of the algorithm. 
For FormulaGPT, during the generation process, every complete expression obtained is counted as an intermediate expression.
For DSO, which samples 500 expressions at a time, we pick the one with the best r2 as an intermediate expression. For SPL, which each pass through MCTS yield a complete expression, we count as obtaining an intermediate expression.

The following table \ref{tab-evaluation} shows the comparison of inference process expressions of FormulaGPT, FormulaGPT without a `shortcut' training set, and two reinforcement learning-based symbolic regression algorithms. 

From the table, it can be clearly seen that formulaGPT's inference process has a smaller number of expressions than DSO, SPL, and FormulaGPT without introducing the `shortcut' training set, so the inference efficiency is higher. This also justifies the introduction of our `shortcut' training dataset.

\begin{table}[ht]
\center
\vspace{-0.2cm}
\caption{The table shows the comparison of the number of expressions in the inference process during the search of the four algorithms.}
\label{tab-evaluation}
\resizebox{8cm}{!}{
\begin{tabular}{ccccc}
 \toprule
 Dataset   & Our         &DSO     & SPL    & Our-No-shortcut \\ 
 \toprule
 Nguyen-1  & 6.2        & 7.9     & 12.3  & 7.8   \\
 Nguyen-2  & 8.2        & 10.5    & 15.2  & 10.6  \\
 Nguyen-3  & 10.2       & 16.3    & 20.4  & 13.6  \\
 Nguyen-4  & 14.1       & 20.6    & 27.2  & 20.3  \\
 Nguyen-5  & 16.4       & 28.0    & 27.4  & 22.4 \\
 Nguyen-6  & 14.2       & 20.7    & 24.8  & 17.3  \\
 Nguyen-7  & 16.0       & 24.3    & 27.3  & 20.8 \\
 Nguyen-8  & 1.2        & 1.1     & 7.2   & 1.3   \\ 
 Nguyen-9  & 15.0       & 28.5    & 25.3  & 19.2  \\ 
 Nguyen-10 & 15.2       & 29.4    & 34.1  & 24.5 \\
 Nguyen-11 & 13.5       & 19.8    & 27.7  & 18.2  \\
 Nguyen-12 & 20.1       & 28.5    & 29.1  & 25.3\\
  \cline{2-5}
 Average   & 12.5       & 19.6    & 23.2  & 16.8  \\

\end{tabular}
}
\end{table}

\section{Appendix: Test data in detail}
Table \ref{a-tab1},\ref{a-tab2},\ref{a-tab3} shows in detail the expression forms of the data set used in the experiment, as well as the sampling range and sampling number. Some specific presentation rules are described below
\begin{itemize}
\item The variables contained in the regression task are represented as [$x_1,x_2,...,x_n$].
\item $U(a,b,c)$ signifies $c$ random points uniformly sampled between $a$ and $b$ for each input variable. Different random seeds are used for training and testing datasets.
\item $E(a,b,c)$ indicates $c$ points evenly spaced between $a$ and $b$ for each input variable. 
\end{itemize}
\begin{table*}[htbp]
\centering
\caption{ Specific formula form and value range of the three data sets Nguyen, Korns, and Jin. 
}
\begin{scriptsize}
\begin{tabular}{ccccc}
\toprule[1.45pt]
\toprule
Name & Expression & Dataset  \\ \hline
Nguyen-1 & $x_1^3+x_1^2+x_1$&U$(-1, 1, 20)$\\
Nguyen-2 & $x_1^4+x_1^3+x_1^2+x_1$ & U$(-1, 1, 20)$ \\
Nguyen-3 & $x_1^5+x_1^4+x_1^3+x_1^2+x_1$ & U$(-1, 1, 20)$ \\
Nguyen-4 & $x_1^6+x_1^5+x_1^4+x_1^3+x_1^2+x_1$ & U$(-1, 1, 20)$  \\
Nguyen-5 & $\sin(x_1^2)\cos(x)-1$ & U$(-1, 1, 20)$  \\
Nguyen-6 & $\sin(x_1)+\sin(x_1+x_1^2)$ & U$(-1, 1, 20)$  \\
Nguyen-7 & $\log(x_1+1)+\log(x_1^2+1)$ & U$(0, 2, 20)$  \\
Nguyen-8 & $\sqrt{x}$ & U$(0, 4, 20)$  \\
Nguyen-9 & $\sin(x)+\sin(x_2^2)$ & U$(0, 1, 20)$ \\
Nguyen-10 & $2\sin(x)\cos(x_2)$ & U$(0, 1, 20)$ \\
Nguyen-11 & $x_1^{x_2}$ & U$(0, 1, 20)$  \\
Nguyen-12 & $x_1^4-x_1^3+\frac{1}{2}x_2^2-x_2$ & U$(0, 1, 20)$ \\
\toprule
Nguyen-2$'$ & $4x_1^4+3x_1^3+2x_1^2+x$ & U$(-1, 1, 20)$  \\
Nguyen-5$'$ & $\sin(x_1^2)\cos(x)-2$ & U$(-1, 1, 20)$  \\
Nguyen-8$'$ & $\sqrt[3]{x}$ & U$(0, 4, 20)$ \\
Nguyen-8$''$ & $\sqrt[3]{x_1^2}$ & U$(0, 4, 20)$ \\
\toprule
Nguyen-1\textsuperscript{c} & $3.39x_1^3+2.12x_1^2+1.78x$ & U$(-1, 1, 20)$ \\
Nguyen-5\textsuperscript{c} & $\sin(x_1^2)\cos(x)-0.75$ & $U(-1, 1, 20)$  \\
Nguyen-7\textsuperscript{c} & $\log(x+1.4)+\log(x_1^2+1.3)$ & U$(0, 2, 20)$ \\
Nguyen-8\textsuperscript{c} & $\sqrt{1.23 x}$ & U$(0, 4, 20)$  \\
Nguyen-10\textsuperscript{c} & $\sin(1.5x)\cos(0.5x_2)$ & U$(0, 1, 20)$  \\
\toprule
Korns-1 & $1.57+24.3*x_1^4$ & U$(-1, 1, 20)$  \\
Korns-2 & $0.23+14.2\frac{(x_4+x_1)}{(3x_2)}$ & U$(-1, 1, 20)$  \\
Korns-3 & $4.9\frac{(x_2-x_1+\frac{x_1}{x_3}}{(3x_3))}-5.41$ & U$(-1, 1, 20)$ \\
Korns-4 & $0.13sin(x_1)-2.3$ & U$(-1, 1, 20)$  \\
Korns-5 & $3+2.13log(|x_5|)$ & U$(-1, 1, 20)$  \\
Korns-6 & $1.3+0.13\sqrt{|x_1|}$ & U$(-1, 1, 20)$  \\
Korns-7 & $2.1(1-e^{-0.55x_1})$ & U$(-1,1 , 20)$  \\
Korns-8 & $6.87+11\sqrt{|7.23 x_1 x_4 x_5|}$ & U$(-1, 1, 20)$ \\
Korns-9 & $12\sqrt{|4.2x_1x_2x_2|}$ & U$(-1, 1, 20)$ \\
Korns-10 & $0.81+24.3\frac{2x_{1}+3x_2^2}{4x_3^3+5x_4^4}$ & U$(-1, 1, 20)$  \\
Korns-11 & $6.87+11cos(7.23x_1^3)$ & U$(-1, 1, 20)$  \\
Korns-12 & $2-2.1cos(9.8x_1^3)sin(1.3x_5)$ & U$(-1, 1, 20)$  \\ 
Korns-13 & $32.0-3.0\frac{tan(x_1)}{tan(x_2)}\frac{tan(x_3)}{tan(x_4)}$ & U$(-1, 1, 20)$ \\
Korns-14 & $22.0-(4.2cos(x_1)-tan(x_2))\frac{tanh(x_3)}{sin(x_4)}$ & U$(-1, 1, 20)$  \\
Korns-15 & $12.0-\frac{6.0tan(x_1)}{e^{x_2}}(log(x_3)-tan(x_4))))$ & U$(-1, 1, 20)$  \\ 
\toprule
Jin-1 & $2.5 x_1^4-1.3 x_1^3 +0.5 x_2^2 - 1.7x_2$ & U$(-3, 3, 100)$ \\
Jin-2 & $8.0 x_1^2 + 8.0 x_2^3 - 15.0$ & U$(-3, 3, 100)$  \\
Jin-3 & $0.2 x_{1}^{3} + 0.5 x_{2}^{3} - 1.2 x_2 - 0.5 x_{1}$ & U$(-3, 3, 100)$  \\    
Jin-4 & $1.5 \exp{x} + 5.0 cos(x_2)$ & U$(-3, 3, 100)$\\
Jin-5 & $6.0 sin(x_1) cos(x_2)$ & U$(-3, 3, 100)$ \\
Jin-6 & $1.35 x_1 x_2 + 5.5 sin((x_1 - 1.0)(x_2 - 1.0))$ & U$(-3, 3, 100)$ \\   
\newline
\end{tabular}
\end{scriptsize}
\label{a-tab1}
\end{table*}

\begin{table*}[htpb]
\centering
\caption{
Specific formula form and value range of the three data sets neat, Keijzer, and Livermore.
}
\begin{scriptsize}
\begin{tabular}{ccccc}
\toprule[1.45pt]
\toprule
Name & Expression & Dataset \\
\hline
Neat-1 & $x_1^4+x_1^3+x_1^2+x$ & U$(-1, 1, 20)$  \\
Neat-2 & $x_1^5+x_1^4+x_1^3+x_1^2+x$ & U$(-1, 1, 20)$ \\
Neat-3 & $\sin(x_1^2)\cos(x)-1$ & U$(-1, 1, 20)$ \\
Neat-4 & $\log(x+1)+\log(x_1^2+1)$ & U$(0, 2, 20)$  \\
Neat-5 & $2\sin(x)\cos(x_2)$ & U$(-1, 1, 100)$  \\
Neat-6 & $\sum_{k=1}^x \frac{1}{k} $ & E$(1, 50, 50)$  \\
Neat-7 & $2 - 2.1\cos(9.8x_1)\sin(1.3x_2)$ & E$(-50, 50, 10^5)$ \\
Neat-8 & $\frac{e^{-(x_1)^2}}{1.2 + (x_2-2.5)^2}$ & U$(0.3, 4, 100)$  \\
Neat-9 & $\frac{1}{1+x_1^{-4}} + \frac{1}{1+x_2^{-4}}$ & E$(-5, 5, 21)$ \\
\toprule
Keijzer-1 & $0.3x_1sin(2\pi x_1)$ & U$(-1, 1, 20)$  \\
Keijzer-2 & $2.0x_1sin(0.5\pi x_1)$ & U$(-1, 1, 20)$  \\
Keijzer-3 & $0.92x_1sin(2.41\pi x_1)$ & U$(-1, 1, 20)$ \\
Keijzer-4 & $x_1^3e^{-x_1}cos(x_1)sin(x_1)sin(x_1)^{2}cos(x_1)-1$ & U$(-1, 1, 20)$ \\
Keijzer-5 & $3+2.13log(|x_5|)$ & U$(-1, 1, 20)$\\

Keijzer-6 & $\frac{x1(x1+1)}{2}$& U$(-1, 1, 20)$ \\
Keijzer-7 & $log(x_1)$ & U$(0,1 , 20)$ \\
Keijzer-8 & $\sqrt{(x_1)}$ & U$(0, 1, 20)$  \\
Keijzer-9 & $log(x_1+\sqrt{x_1^2}+1)$ & U$(-1, 1, 20)$ \\
Keijzer-10 & $x_{1}^{x_2}$ & U$(-1, 1, 20)$  \\
Keijzer-11 & $x_1x_2+sin((x_1-1)(x_2-1))$ & U$(-1, 1, 20)$  \\
Keijzer-12 & $x_1^4-x_1^3+\frac{x_2^2}{2}-x_2$ & U$(-1, 1, 20)$  \\ 
Keijzer-13 & $6sin(x_1)cos(x_2)$ & U$(-1, 1, 20)$  \\
Keijzer-14 & $\frac{8}{2+x_1^2 + x_2^2}$ & U$(-1, 1, 20)$ \\
Keijzer-15 & $\frac{x_1^3}{5}+\frac{x_2^3}{2}-x_2-x_1$ & U$(-1, 1, 20)$ \\ 

\toprule
Livermore-1 & $\frac{1}{3}+x_1+sin(x_1^2))$ & U$(-3, 3, 100)$  \\
Livermore-2 & $sin(x_1^2)*cos(x1)-2$ & U$(-3, 3, 100)$  \\
Livermore-3 & $sin(x_1^3)*cos(x_1^2))-1$ & U$(-3, 3, 100)$  \\
Livermore-4 & $log(x_1+1)+log(x_1^2+1)+log(x_1)$ & U$(-3, 3, 100)$ \\ 
Livermore-5 & $x_1^4-x_1^3+x_2^2-x_2$ & U$(-3, 3, 100)$  \\
Livermore-6 & $4x_1^4+3x_1^3+2x_1^2+x_1$ & U$(-3, 3, 100)$ \\ 
Livermore-7 & $\frac{(exp(x1)-exp(-x_1)}{2})$ & U$(-1, 1, 100)$ \\ 
Livermore-8 & $\frac{(exp(x1)+exp(-x1)}{3}$ & U$(-3, 3, 100)$ \\
Livermore-9 & $x_1^9+x_1^8+x_1^7+x_1^6+x_1^5+x_1^4+x_1^3+x_1^2+x_1$ & U$(-1, 1, 100)$  \\
Livermore-10 & $6*sin(x_1)cos(x_2)$ & U$(-3, 3, 100)$  \\
Livermore-11 & $\frac{x_1^2 x_2^2}{(x_1+x_2)}$ & U$(-3, 3, 100)$ \\
Livermore-12 & $\frac{x_1^5}{x_2^3}$ & U$(-3, 3, 100)$  \\
Livermore-13 & $x_1^{\frac{1}{3}}$ & U$(-3, 3, 100)$  \\
Livermore-14 & $x_1^3+x_1^2+x_1+sin(x_1)+sin(x_2^2)$ & U$(-1, 1, 100)$ \\ 
Livermore-15 & $x_1^\frac{1}{5}$ & U$(-3, 3, 100)$  \\
Livermore-16 & $x_1^{\frac{2}{3}}$ & U$(-3, 3, 100)$  \\  
Livermore-17 & $4sin(x_1)cos(x_2)$ & U$(-3, 3, 100)$  \\
Livermore-18 & $sin(x_1^2)*cos(x_1)-5$ & U$(-3, 3, 100)$  \\
Livermore-19 & $x_1^5+x_1^4+x_1^2 + x_1$ & U$(-3, 3, 100)$  \\
Livermore-20 & $e^{(-x_1^2)}$ & U$(-3, 3, 100)$  \\
Livermore-21 & $x_1^8+x_1^7+x_1^6+x_1^5+x_1^4+x_1^3+x_1^2+x_1$& U$(-1, 1, 20)$ \\
Livermore-22 & $e^{(-0.5x_1^2)}$ & U$(-3, 3, 100)$  \\
\newline
\end{tabular}
\end{scriptsize}
\label{a-tab2}
\end{table*}

\begin{table*}[htpb]
\centering
\caption{
Specific formula form and value range of the three data sets Vladislavleva and others. }
\begin{scriptsize}
\begin{tabular}{ccccc}
\toprule[1.45pt]
\toprule
Name & Expression & Dataset \\
\toprule
Vladislavleva-1 & $\frac{(e^{-(x1-1)^2})}{(1.2+(x2-2.5)^2))}$ & U$(-1, 1, 20)$ \\
Vladislavleva-2 & $e^{-x_1}x_1^3cos(x_1)sin(x_1)(cos(x_1)sin(x_1)^2-1)$ & U$(-1, 1, 20)$ \\

Vladislavleva-3 & $e^{-x_1}x_1^3cos(x_1)sin(x_1)(cos(x_1)sin(x_1)^2-1)(x_2-5)$ & U$(-1, 1, 20)$ \\
Vladislavleva-4 & $\frac{10}{5+(x1-3)^2+(x_2-3)^2+(x_3-3)^2+(x_4-3)^2+(x_5-3)^2}$ & U$(0, 2, 20)$ \\
Vladislavleva-5 & $30(x_1-1)\frac{x_3-1}{(x_1-10)}x_2^2$ & U$(-1, 1, 100)$ \\
Vladislavleva-6 & $6sin(x_1)cos(x_2)$ & E$(1, 50, 50)$ \\
Vladislavleva-7 & $2 - 2.1\cos(9.8x)\sin(1.3x_2)$ & E$(-50, 50, 10^5)$ \\
Vladislavleva-8 & $\frac{e^{-(x-1)^2}}{1.2 + (x_2-2.5)^2}$ & U$(0.3, 4, 100)$  \\
\toprule
Test-2 & $3.14x_1^2$ & U$(-1, 1, 20)$ \\
Const-Test-1 & $5x_1^2$ & U$(-1, 1, 20)$ \\
GrammarVAE-1 & $1/3+x1+sin(x_1^2))$ & U$(-1, 1, 20)$ \\
Sine & $sin(x_1)+sin(x_1+x_1^2))$ & U$(-1, 1, 20)$ \\
Nonic & $x_1^9+x_1^8+x_1^7+x_1^6+x_1^5+x_1^4+x_1^3+x_1^2+x_1$ & U$(-1, 1, 100)$  \\
Pagie-1 & $\frac{1}{1+x_1^{-4}+\frac{1}{1+x2^{-4}}} $ & E$(1, 50, 50)$  \\
Meier-3 & $\frac{x_1^2  x_2^2}{(x_1+x_2)}$ & E$(-50, 50, 10^5)$ \\
Meier-4 & $\frac{x_1^5}{x_2^3}$ & $U(0.3, 4, 100)$  \\
Poly-10 & $x_1x_2+x_3x4+x_5x_6+x_1x_7x_9+x_3x_6x_{10}$ & E$(-1, 1, 100)$ \\
\toprule
Constant-1 & $3.39*x_1^3+2.12*x_1^2+1.78*x_1$&$U(-4, 4, 100)$\\
Constant-2 & $sin(x_1^2)*cos(x_1)-0.75$&$U(-4, 4, 100)$\\
Constant-3 & $sin(1.5*x_1)*cos(0.5*x_2)$&$U(0.1, 4, 100)$\\
Constant-4 & $2.7*x_1^{x_2}$&$U(0.3, 4, 100)$\\
Constant-5 & $sqrt(1.23*x_1)$&$U(0.1, 4, 100)$\\
Constant-6 & $x_1^{0.426}$&$U(0.0, 4, 100)$\\
Constant-7 & $2*sin(1.3*x_1)*cos(x_2)$&$U(-4, 4, 100)$\\
Constant-8 & $log(x_1+1.4)+log(x1,2+1.3)$&$U(-4, 4, 100)$\\
\toprule
R1 & $\frac{(x_1+1)^3}{x_1^2-x_1+1)}$&$U(-5, 5, 100)$\\
R2 & $\frac{(x_1^2-3*x_1^2+1}{x_1^2+1)}$&$U(-4, 4, 100)$\\
R3 & $\frac{x_1^6+x_1^5)}{(x_1^4+x_1^3+x_1^2+x1+1)}$&$U(-4, 4, 100)$\\
\newline
\end{tabular}
\end{scriptsize}
\label{a-tab3}
\end{table*}

\section{Appendix: FormulaGPT tests on AIFeynman dataset.}

In our study, we conducted an evaluation of our novel symbol regression algorithm, termed FormulaGPT, leveraging the AI Feynman dataset, which comprises a diverse array of problems spanning various subfields of physics and mathematics, including mechanics, thermodynamics, and electromagnetism. Originally, the dataset contained 100,000 data points; however, for a more rigorous assessment of FormulaGPT's efficacy, our analysis was deliberately confined to a subset of 100 data points. Through the application of FormulaGPT for symbol regression on these selected data points, we meticulously calculated the $R^2$ values to compare the algorithm's predictions against the true solutions.

The empirical results from our investigation unequivocally affirm that FormulaGPT possesses an exceptional ability to discern the underlying mathematical expressions from a constrained sample size. Notably, the $R^2$ values achieved were above 0.99 for a predominant portion of the equations, underscoring the algorithm's remarkable accuracy in fitting these expressions. These findings decisively position FormulaGPT as a potent tool for addressing complex problems within the domains of physics and mathematics. The broader implications of our study suggest that FormulaGPT holds considerable promise for a wide range of applications across different fields. Detailed experimental results are presented in Table \ref{a-tab5} and Table \ref{a-tab6}.

\begin{table}[htbp]
\centering
{\footnotesize
\begin{tabular}{|l|l|r|}
\hline
Feynman   & Equation & $R^2$ \\
\hline                            
I.6.20a       & $f = e^{-\theta^2/2}/\sqrt{2\pi}$ & 0.9999  \\
I.6.20        & $f = e^{-\frac{\theta^2}{2\sigma^2}}/\sqrt{2\pi\sigma^2}$ & 0.9983\\
I.6.20b       & $f = e^{-\frac{(\theta-\theta_1)^2}{2\sigma^2}}/\sqrt{2\pi\sigma^2}$ & 0.9934 \\
I.8.14       & $d = \sqrt{(x_2-x_1)^2+(y_2-y_1)^2}$ & 0.9413  \\
I.9.18       & $F = \frac{Gm_1m_2}{(x_2-x_1)^2+(y_2-y_1)^2+(z_2-z_1)^2}$  & 0.9938\\
I.10.7       & $F = \frac{Gm_1m_2}{(x_2-x_1)^2+(y_2-y_1)^2+(z_2-z_1)^2}$  & 0.9782\\
I.11.19      & $A = x_1y_1+x_2y_2+x_3y_3$ & 0.9993   \\
I.12.1       & $F = \mu N_n$ & 0.9999 \\
I.12.2       & $F = \frac{q_1q_2}{4\pi\epsilon r^2}$   & 0.9999 \\
I.12.4       & $E_f = \frac{q_1}{4\pi\epsilon r^2}$  & 0.9999 \\
I.12.5       & $F = q_2 E_f$ & 0.9999  \\
I.12.11      & $F = \mathcal{Q}(E_f+B v \sin\theta)$  & 0.9969 \\
I.13.4      & $K = \frac{1}{2}m(v^2+u^2+w^2)$  & 0.9831  \\
I.13.12      & $U = Gm_1m_2(\frac{1}{r_2}-\frac{1}{r_1})$ & 0.9999  \\
I.14.3       & $U = mgz$ &1.0    \\
I.14.4       & $U = \frac{k_{spring}x^2}{2}$  & 0.9924  \\
I.15.3x      & $x_1 = \frac{x-ut}{\sqrt{1-u^2/c^2}}$ & 0.9815 \\
I.15.3t      & $t_1 = \frac{t-ux/c^2}{\sqrt{1-u^2/c^2}}$ & 0.9822  \\
I.15.10       & $p = \frac{m_0v}{\sqrt{1-v^2/c^2}}$ & 0.9920 \\
I.16.6       & $v_1 = \frac{u+v}{1+uv/c^2}$ & 0.9903  \\
I.18.4       & $r = \frac{m_1r_1+m_2r_2}{m_1+m_2}$ & 0.9711 \\
I.18.12      & $\tau = rF\sin\theta$  & 0.9999  \\
I.18.16      & $L = mrv \sin\theta$  & 0.9997 \\
I.24.6 & $E = \frac{1}{4} m (\omega^2+\omega_0^2) x^2$      & 0.9991\\
I.25.13      & $V_e = \frac{q}{C}$ & 1.0 \\
I.26.2       & $\theta_1 = \arcsin(n  \sin\theta_2)$ & 0.9989 \\
I.27.6       & $f_f$    $ = \frac{1}{\frac{1}{d_1}+\frac{n}{d_2}}$  & 0.9942 \\
I.29.4       & $k = \frac{\omega}{c}$ & 1.0 \\
I.29.16      & $x = \sqrt{x_1^2+x_2^2-2x_1x_2\cos(\theta_1-\theta_2)}$ & 0.9922  \\
I.30.3 & $I_* = I_{*_0}\frac{\sin^2(n\theta/2)}{\sin^2(\theta/2)}$ & 0.9946 \\
I.30.5       & $\theta = \arcsin(\frac{\lambda}{nd})$  & 0.9933\\
I.32.5       & $P = \frac{q^2a^2}{6\pi\epsilon c^3}$       & 0.9905 \\
I.32.17 & $P = (\frac{1}{2}\epsilon c E_f^2)(8\pi r^2/3) (\omega^4/(\omega^2-\omega_0^2)^2)$      & 0.9941  \\
I.34.8       & $\omega = \frac{qvB}{p}$   & 1.0\\
I.34.10       & $\omega = \frac{\omega_0}{1-v/c}$ & 0.9903 \\
I.34.14      & $\omega = \frac{1+v/c}{\sqrt{1-v^2/c^2}}\omega_0$  & 0.9941 \\
I.34.27      & $E = \hbar\omega$  & 0.9999 \\
I.37.4       & $I_* = I_1+I_2+2\sqrt{I_1I_2}\cos\delta$ & 0.9723\\
I.38.12      & $r = \frac{4\pi\epsilon\hbar^2}{mq^2}$   & 0.9999  \\
I.39.10       & $E = \frac{3}{2}p_F V$     & 0.9981 \\
I.39.11      & $E = \frac{1}{\gamma-1}p_F V$  & 0.9883 \\
I.39.22      & $P_F = \frac{n k_b T}{V}$       & 0.9902  \\
I.40.1       & $n = n_0e^{-\frac{mgx}{k_bT}}$    & 0.9924 \\
I.41.16      & $L_{rad} = \frac{\hbar\omega^3}{\pi^2c^2(e^{\frac{\hbar\omega}{k_bT}}-1)}$ & 0.9435  \\
I.43.16      & $v = \frac{\mu_{drift}q V_e}{d}$   & 0.9903  \\
I.43.31      & $D = \mu_e k_bT$    & 1.0  \\
I.43.43      & $\kappa = \frac{1}{\gamma-1}\frac{k_bv}
{A}$  & 0.9333  \\
I.44.4       & $E = n k_b T \ln(\frac{V_2}{V_1})$   & 0.8624  \\
I.47.23      & $c = \sqrt{\frac{\gamma pr}{\rho}}$   & 0.9624 \\
I.48.20       & $E = \frac{m c^2}{\sqrt{1-v^2/c^2}}$ &  0.8866\\
I.50.26 & $x = x_1[\cos(\omega t)+\alpha\> cos(\omega t)^2]$      & 0.9999   \\
\hline
\end{tabular}
\caption{Tested Feynman Equations, part 1.}
\label{a-tab5}
}
\end{table}
\begin{table*}[htbp]
\centering
{\footnotesize

\begin{tabular}{|l|l|r|}
\hline
Feynman   & Equation & $R^2$\\
\hline       
II.2.42   & P     $ = \frac{\kappa(T_2-T_1)A}{d}$  & 0.8335  \\
II.3.24   & $F_E = \frac{P}{4\pi r^2}$  & 0.9755 \\
II.4.23   & $V_e = \frac{q}{4\pi\epsilon r}$   & 0.9901 \\
II.6.11 & $V_e =\frac{1}{4\pi\epsilon}\frac{p_d\cos \theta}{r^2}$      & 0.9913 \\
II.6.15a & $E_f = \frac{3}{4\pi\epsilon}\frac{p_d z}{r^5} \sqrt{x^2+y^2}$      & 0.9031  \\
II.6.15b & $E_f = \frac{3}{4\pi\epsilon}\frac{p_d}{r^3} \cos\theta\sin\theta$      & 0.9925  \\
II.8.7    & $E = \frac{3}{5}\frac{q^2}{4\pi\epsilon d}$  & 0.9736  \\
II.8.31   & $E_{den} = \frac{\epsilon E_f^2}{2}$                     & 0.9999 \\
II.10.9   & $E_f = \frac{\sigma_{den}}{\epsilon}\frac{1}{1+\chi}$      & 0.9939  \\
II.11.3 & $x = \frac{q E_f}{m(\omega_0^2-\omega^2)}$      & 0.9903     \\
II.11.7 & $n = n_0(1+ \frac{p_d E_f \cos\theta}{k_b T})$      & 0.8305 \\
II.11.20  & $P_* = \frac{n_\rho p_d^2 E_f}{3 k_b T}$ & 0.8013  \\
II.11.27 & $P_* = \frac{n\alpha}{1-n\alpha/3}\epsilon E_f$      & 0.9816   \\
II.11.28  & $\theta = 1+\frac{n\alpha}{1-(n\alpha/3)}$    & 0.9985\\ 
II.13.17  & $B = \frac{1}{4 \pi \epsilon c^2}\frac{2I}{r}$ & 0.9991\\
II.13.23  & $\rho_c = \frac{\rho_{c_0}}{\sqrt{1-v^2/c^2}}$          & 0.9832  \\
II.13.34  & $j = \frac{\rho_{c_0}v}{\sqrt{1-v^2/c^2}}$     & 0.9747 \\
II.15.4   & $E = -\mu_M B \cos\theta$               & 0.9999 \\
II.15.5   & $E = -p_d E_f\cos\theta$  & 0.9964 \\
II.21.32  & $V_e = \frac{q}{4\pi\epsilon r(1-v/c)}$   & 0.9899   \\
II.24.17 & $k = \sqrt{\frac{\omega^2}{c^2}-\frac{\pi^2}{d^2}}$      & 0.9835   \\
II.27.16  & $F_E = \epsilon c E_f^2$        & 0.9954 \\
II.27.18  & $E_{den} = \epsilon E_f^2$         & 0.9952 \\
II.34.2a  & $I = \frac{qv}{2\pi r}$         & 0.9835 \\
II.34.2   & $\mu_M = \frac{q v r}{2}$             & 0.9946 \\
II.34.11  & $\omega = \frac{g_{\_} q B}{2m}$          & 0.9935 \\
II.34.29a & $\mu_M = \frac{q h}{4\pi m}$      & 0.9956  \\
II.34.29b & $E = \frac{g_{\_} \mu_M B J_z}{\hbar}$ & 0.8614\\
II.35.18 & $n = \frac{n_0}{\exp(\mu_m B/(k_b T))+\exp(-\mu_m B/(k_b T))}$      & 0.9647 \\
II.35.21  & $M = n_\rho \mu_M \tanh(\frac{\mu_M B}{k_b T})$     & 0.8097 \\
II.36.38 & $f = \frac{\mu_m B}{k_b T}+\frac{\mu_m\alpha M}{\epsilon c^2 k_b T}$      & 0.9840\\
II.37.1   & $E = \mu_M(1+\chi)B$    & 0.9999\\
II.38.3   & $F = \frac{Y A x}{d}$            & 0.9979 \\
II.38.14  & $\mu_S = \frac{Y}{2(1+\sigma)}$     & 0.9999  \\
III.4.32  & $n = \frac{1}{e^{\frac{\hbar\omega}{k_bT}}-1}$ & 0.9812  \\
III.4.33  & $E = \frac{\hbar\omega}{e^{\frac{\hbar\omega}{k_b T}}-1}$  & 0.9964    \\
III.7.38  & $\omega = \frac{2 \mu_M B}{\hbar}$  & 0.9932  \\
III.8.54  & $p_{\gamma}$    $ = \sin(\frac{E t}{\hbar})^2$  & 0.9943\\
III.9.52  & $p_{\gamma}$    $ = \frac{p_d E_f t}{\hbar} \frac{    \sin((\omega-\omega_0)t/2)^2}{((\omega-\omega_0)t/2)^2}$ & 0.7622  \\
III.10.19 & $E = \mu_M\sqrt{B_x^2+B_y^2+B_z^2}$  & 0.9964 \\
III.12.43 & $L = n\hbar$ & 0.9993  \\
III.13.18 & $v = \frac{2 E d^2 k}{\hbar}$ & 0.9959  \\
III.14.14 & $I = I_0 (e^{\frac{q V_e}{k_b T}}-1)$  & 0.9925\\
III.15.12 & $E = 2U(1-\cos(kd))$    & 0.9998 \\
III.15.14 & $m = \frac{\hbar^2}{2E d^2}$     & 0.9947  \\
III.15.27 & $k = \frac{2\pi\alpha}{nd}$    & 0.9992 \\
III.17.37 & $f = \beta(1+\alpha \cos\theta)$ & 0.9925 \\
III.19.51 & $E = \frac{-mq^4}{2(4\pi\epsilon)^2\hbar^2}\frac{1}{n^2}$     & 0.9934 \\
III.21.20 & $j = \frac{-\rho_{c_0} q A_{vec}}{m}$  & 0.8036  \\
\hline
\end{tabular} 
\caption{Tested Feynman Equations, part 2.}
\label{a-tab6}
}
\end{table*}

\end{document}